\documentclass[Journal,letterpaper,NoLineNumbers]{ascelike-new}
\usepackage[utf8]{inputenc}
\usepackage[T1]{fontenc}
\usepackage{lmodern}
\usepackage{graphicx}
\usepackage[style=base,figurename=Fig.,labelfont=bf,labelsep=period]{caption}
\usepackage{subcaption}
\usepackage{amsmath}
\usepackage{newtxtext,newtxmath}
\usepackage[colorlinks=true,citecolor=red,linkcolor=black]{hyperref}
\usepackage{booktabs}
\usepackage{multirow}
\usepackage{tcolorbox}
\tcbuselibrary{listings, breakable}
\usepackage{listings}
\usepackage{amsthm}
\theoremstyle{definition}
\newtheorem{definition}{Definition}
\NameTag{Zhang, \today}
\begin{document}

\title{Automated Facility Enumeration for Building Compliance Checking using Door Detection and Large Language Models}

\author[a]{Licheng Zhang\thanks{Corresponding author.}}
\author[a]{Bach Le}
\author[a]{Naveed Akhtar}
\author[a]{Tuan Ngo}

\affil[a]{The University of Melbourne, Parkville, Victoria 3010, Australia. Email: licheng.zhang@student.unimelb.edu.au}

\maketitle

\begin{abstract}
Building compliance checking (BCC) is a critical process for ensuring that constructed facilities meet regulatory standards. A core component of BCC is the accurate enumeration of facility types and their spatial distribution. Despite its importance, this problem has been largely overlooked in the literature, posing a significant challenge for BCC and leaving a critical gap in existing workflows. Performing this task manually is time-consuming and labor-intensive. Recent advances in large language models (LLMs) offer new opportunities to enhance automation by combining visual recognition with reasoning capabilities. In this paper, we introduce a new task for BCC: automated facility enumeration, which involves validating the quantity of each facility type against statutory requirements. To address it, we propose a novel method that integrates door detection with LLM-based reasoning. We are the first to apply LLMs to this task and further enhance their performance through a Chain-of-Thought (CoT) pipeline. Our approach generalizes well across diverse datasets and facility types. Experiments on both real-world and synthetic floor plan data demonstrate the effectiveness and robustness of our method.
\end{abstract}

\section{Practical Applications}
This work demonstrates the potential of LLMs to achieve accurate and generalizable automated facility enumeration. By leveraging LLMs’ capability to interpret design layouts and reason over facility placements, the proposed method enables automatic verification of sanitary facilities, kitchens, laundries, exits, fire safety equipment, accessibility provisions, and parking with minimal human intervention. It supports practitioners by accelerating design validation, reducing manual inspection efforts, and enhancing the accuracy of facility enumeration across diverse building projects.
\section{Author keywords}
\noindent Building compliance checking; Facility enumeration; Door detection; Large language models; Chain-of-Thought reasoning.
\section{Introduction}
Building compliance checking (BCC) has emerged as a critical area of research in the architecture, engineering, and construction (AEC) domain due to its direct implications for safety, functionality, and sustainability in the built environment. In Australia, compliance is governed by the National Construction Code (NCC) \cite{abcb2022ncc}, the nation’s primary regulatory framework developed by the Australian Building Codes Board and enforced through state and territory legislation. The NCC specifies minimum standards for building design and construction to ensure safety, health, amenity, accessibility, and environmental performance. Organized into three volumes, it covers a wide range of building classes, including residential, commercial, industrial, educational, and public facilities. Compliance with the NCC is mandatory, making it an indispensable reference for regulating facility provision and distribution. A core component of BCC under the NCC is facility enumeration, which calls for a systematic verification that sanitary facilities, kitchens, laundries, exits, fire safety equipment, accessibility features, and parking provisions etc.~meet statutory requirements.

Traditionally, building compliance has been verified through manual checking \cite{zhang2022towards,villaschi2022bim}, a process that is both time-consuming and labor-intensive. Automated solutions, such as object detection \cite{yolov8}, have been investigated to alleviate this burden \cite{zaidi2024vision}. However, its effectiveness remains limited due to poor generalization, as visual objects vary considerably across datasets, architectural conventions, and domains, leading to inconsistent performance when applied outside the training distribution \cite{wu2022model}.

Recent advances in Large Language Models (LLMs) \cite{hurst2024gpt,openai2025gpt5} open new opportunities for automating BCC. Unlike traditional vision-only approaches, LLMs integrate multimodal reasoning by combining visual understanding with textual interpretation \cite{dong2025insight}. Trained on massive and diverse datasets, they possess the ability to generalize across regulatory contexts, interpret complex building requirements, and jointly reason over images and text. These capabilities hold strong potential for more intelligent and scalable compliance checking workflows, reducing manual workload while improving accuracy, robustness, and adaptability \cite{chen2025vision}. Although still at an early stage, LLM-based approaches show promise as intelligent assistants for the AEC industry, with the potential to significantly transform compliance verification practices \cite{katooziani2025gpt}.

Within this context, CAD floor plan drawings represent an ideal medium for BCC. As visual abstractions of architectural layouts, CAD drawings are widely available, easily interpretable, and more lightweight than structured alternatives such as building information modeling (BIM). Their visual clarity makes them particularly suitable for automated analysis of regulatory adherence.

Leveraging the above facts, this work introduces a new task of \textit{automated facility enumeration} from CAD floor plan images, which serves as a critical step toward scalable and accurate BCC under the NCC. This task goes beyond merely detecting facilities, requiring verification of their quantity, spatial distribution, and compliance with regulatory standards. It is a high-impact and non-trivial problem in intelligent compliance automation, ensuring that all essential facilities, such as sanitary fixtures, kitchens, laundries, exits, fire safety equipment, accessibility features, and parking, are correctly provided, distributed, and configured to meet safety and regulatory requirements. Errors in facility enumeration can lead to serious safety and legal risks. For instance, a missing exit door or fire extinguisher can invalidate a design and delay project approval, while overestimating the number of emergency exits or fire extinguishers may also create safety hazards, such as improper evacuation planning or misallocation of resources. Regulatory agencies mandate strict compliance, and even a single miscount may be unacceptable. Beyond safety, accurate facility enumeration is critical for downstream BCC tasks such as occupant load analysis, fire evacuation modeling, and accessibility audits, and it can be integrated with BIM or digital twin workflows to support end-to-end automated compliance. From an economic and sustainability perspective, automating facility enumeration reduces costly rework, shortens design cycles, and minimizes resource waste caused by late-stage design changes.

At the same time, the task poses significant technical and research challenges. Floor plans are inherently ambiguous, as facility symbols can vary due to differing architectural standards, regional practices, and design styles, and may be partially occluded or overlap with other elements. Facility enumeration involves more than simply counting symbols: some facilities are shared across multiple dwellings, while others depend on occupant load or functional use requiring careful contextual reasoning to ensure accurate compliance verification. This task requires reasoning over both spatial context and regulatory rules. Furthermore, it spans multiple scales, from local features such as doors and sanitary fixtures to global facilities such as site-wide car parking. Each facility type presents heterogeneous representation challenges, rendering the task inherently multi-domain and multi-class. Critically, automated facility enumeration is a zero-tolerance task: overestimating or underestimating even a single instance constitutes non-compliance, distinguishing it from conventional counting or detection problems.

Given these challenges, we propose an approach that leverages the combined visual and textual reasoning capabilities of LLMs, augmented with an intermediate step of door detection. The pipeline begins by detecting door symbols, which are relatively consistent across floor plan drawings and generalize well across domains. Detected doors serve as anchor points for the LLM to perform facility enumeration. Doors are selected for three main reasons: (1) unlike rooms, which vary widely in shape, size, and spatial configuration, doors exhibit stable symbolic representations across drawings, enabling reliable cross-domain detection; (2) they allow the LLM to focus on local regions around each door rather than analyzing the entire image, simplifying reasoning and improving predictive accuracy; and (3) highlighted doors provide reference points for identifying additional instances of the same facility type, such as emergency exits, where visually similar doors can be considered valid instances.

Viewed more broadly, automated facility enumeration can be framed as a multi-modal reasoning benchmark for compliance-critical verification, encompassing the symbol, room, and building levels, together with rule-based validation against NCC standards. Its zero-tolerance requirement across diverse building types makes it uniquely challenging and positions it as a meaningful and non-trivial problem in intelligent, scalable BCC.

Furthermore, motivated by the growing evidence that LLMs achieve superior performance on reasoning-intensive tasks when supported by Chain-of-Thought (CoT) prompting \cite{kojima2022large,wei2022chain}, we propose a CoT-based strategy tailored to facility enumeration. This design leverages intermediate reasoning steps to enhance interpretability and accuracy, positioning our approach as a principled framework for advancing automated BCC.

In summary, the contributions of this paper are as follows:
\begin{itemize}
    \item We introduce a novel task of automated facility enumeration from CAD floor plan images, representing a key step toward scalable automation in BCC.
    \item We propose a new approach that combines door detection with the capabilities of LLMs, incorporating Chain-of-Thought reasoning to perform accurate and generalizable facility enumeration.
    \item We conduct comprehensive experiments across diverse compliance scenarios to validate the effectiveness, robustness, and generalizability of our method.
\end{itemize}
\section{Background}
Under the NCC \cite{abcb2022ncc}, buildings in Australia are categorised into nine classes according to their primary use and occupancy type. \textbf{Class~1}: single residential dwellings, \textbf{Class~2}: apartments, \textbf{Class~3}: hotels, motels, and boarding houses, \textbf{Class~4}:  dwelling within another building (e.g., caretaker’s residence), \textbf{Class~5}: office buildings, \textbf{Class~6}: shops and retail premises, \textbf{Class~7}: carparks and storage buildings, \textbf{Class~8}: factories and laboratories, and \textbf{Class~9}: public buildings such as schools, hospitals, and assembly spaces.  

To ensure compliance with the NCC, it is essential to enumerate building facilities accurately. The requirements vary by facility type across the different building classes, as summarised below. 

\noindent\textbf{Sanitary facilities.} For residential classes (Class~1–4), each dwelling or unit must include at least one toilet, wash basin, and shower or bath. In Class~3 facilities (e.g., hotels), one set of sanitary facilities must be provided for every ten residents without private amenities. For Class~5–9 buildings (offices, shops, factories, public buildings), the number of toilets and wash basins is determined by occupant load. In all non-residential classes, at least one accessible unisex sanitary facility per floor is required.  

\noindent\textbf{Kitchens.} Dwellings (Class~1, 2, 4) must have a functional kitchen. Class~3 facilities require kitchens or kitchenettes for long-term accommodation, while short-term stays may provide shared facilities. For Class~5–9 buildings, kitchens are not always mandatory; they are required only where building use demands them (e.g., staff facilities in offices and factories, canteens in schools, hospital kitchens). Carparks (Class~7) do not require kitchens.  

\noindent\textbf{Laundries.} Residential classes (Class~1, 2, 4) require either an in-unit laundry or access to a shared laundry. Class~3 buildings must provide shared laundry facilities for residents. Laundries are not generally required in Class~5–9 buildings, except where the specific use demands them (e.g., hospitals or aged care in Class~9).  

\noindent\textbf{Exits and egress.} Class~1 dwellings require at least one exit, whereas all other classes (Class~2–9) generally require a minimum of two exits per floor. Multi-storey buildings must include staircases, and ramps are required wherever accessibility is needed.  

\noindent\textbf{Fire and safety equipment.} Smoke alarms are mandatory in all habitable rooms across residential classes (Class~1–4) and in most spaces for Class~5–9 buildings. Non-residential buildings also require fire extinguishers, hose reels, and in many cases sprinklers, with specifications tied to occupant load and assessed risk.  

\noindent\textbf{Accessibility.} Each floor of Class 2–9 buildings must provide at least one accessible entrance and one accessible facility to ensure compliance with accessibility standards. Residential Class~1 dwellings are exempt from common-area accessibility requirements but must comply with smoke alarm provisions. 

\noindent\textbf{Parking.} Class 1 and Class 2 dwellings are required to provide at least one car parking space per dwelling. Additionally, in any building class where parking is provided, at least one accessible parking space must be provided for every fifty standard parking spaces. These requirements represent the minimum mandatory provisions to ensure adequate parking and accessibility.

These requirements establish the regulatory framework for facility enumeration in compliance checking, underscoring the need for automated, scalable methods capable of handling diverse facility types across building classes.  
\section{Related Work}
Owing to our contribution in automation for BCC using LLMs \cite{hettiarachchi2025code,chen2025multi,saluz2025large,adil2025using}, we contextualize our work by focusing on the existing strategies to integrate LLMs with reasoning, prompting, BIM, and vision-language capabilities to improve automated compliance checking.
\subsection{LLMs with Knowledge Representation and Reasoning}
Several studies have focused on combining LLMs with structure knowledge representations to enhance reasoning. For instance, \citeN{fuchs2024using} explored using LLMs to translate complex building regulations into a computable LegalRuleML format, enabling more efficient and automated compliance checking through formalized rule representation. Along the same lines, \citeN{li2024intelligent} developed an intelligent compliance checking method that combined knowledge graphs and LLMs to automatically parse and evaluate construction schemes with enhanced accuracy. Moreover, \citeN{al2024human} introduced a human-in-the-loop framework using GPT-4o \cite{hurst2024gpt} and active learning to automate RASE tagging of building regulations, which improved the accuracy of structured compliance data generation by combining few-shot learning, fine-tuning, and expert feedback. \citeN{hettiarachchi2025code} further contributed by developing a dataset of annotated building regulation sentences designed to facilitate machine-readable rule generation, enabling LLMs to perform entity recognition and relation extraction more effectively in the AEC domain.
\subsection{Prompting, RAG, and Autonomous Agents}
Beyond knowledge representation, other studies have emphasized prompting, retrieval-augmented generation (RAG), and autonomous agent design for leveraging LLMs in AEC. For example, \citeN{liu2023gpt} leveraged prompt engineering in a GPT-based automated compliance checking method to input building design specifications and codes directly, facilitating automated evaluation without additional domain knowledge. In a similar vein, \citeN{yang2024prompt} designed a prompt-based framework using GPT to automate the transformation of building code information into machine-readable formats, significantly reducing manual effort and training data while maintaining high accuracy. \citeN{zhu2024refining} further demonstrated how fine-tuning pre-trained LLMs can improve accuracy and efficiency in reviewing bridge construction compliance. Similarly, \citeN{wang2025llm} introduced HSE-Bench, a benchmark dataset to evaluate LLMs in health, safety, and environment compliance, coupled with a reasoning-of-experts prompting method to improve domain-specific legal reasoning. \citeN{he2025enriched} extended this concept by proposing a hybrid RAG framework combining keyword and semantic search to enhance LLM responses for construction regulation inquiries, thus enabling precise, domain-aware question answering on fragmented and unstructured regulations.  Likewise, \citeN{katooziani2025gpt} systematically evaluated GPT for construction safety management, highlighting its strengths and limitations in automation, critical reasoning, and adaptability. Extending this idea, \citeN{sun2025compliance} proposed an RAG framework that combined LLMs with logic reasoning and semantic embedding techniques using event-centric graphs to improve text-based compliance checking. Finally, \citeN{chen2025multi} proposed a multi-agent LLM framework that decomposed code-compliant reinforced concrete design into specialized LLM agents which coordinated, cross-checked results, and generated verifiable design calculations.
\subsection{LLMs Integrated with BIM}
Integration of LLMs with BIM systems has also been widely explored. For example, \citeN{ying2024automatic} proposed an autonomous compliance checking framework leveraging LLMs as intelligent agents capable of understanding building design requirements, planning verification tasks, retrieving BIM data, and executing checks independently. Similarly, \citeN{chen2024automated} proposed an automated BIM compliance checking framework that integrated LLMs, deep learning, and ontology-based reasoning. In addition, \citeN{nakhaee2024vision} presented a unified framework combining hybrid knowledge graphs and LLMs to automate BCC, enabling semantic reasoning over regulations and BIM data for improved accuracy and scalability. Likewise, \citeN{madireddy2025large} proposed a semi-automated system that integrated LLMs with BIM software to interpret building codes and generate executable scripts for automatic BCC, improving accuracy and efficiency. Furthermore, \citeN{saluz2025large} proposed a CoT-based approach using LLMs to align BIM authoring models with compliance constraints, reducing expert involvement and enabling automated checking for modular Kit-of-Parts architecture.
\subsection{VLMs for Compliance and Safety}
Finally, Vision-Language Models (VLMs) \cite{openai2025gpt41} have also been used for visual compliance and safety monitoring. For instance, \citeN{chen2025vision} proposed a VLM approach using a vision Transformer combined with multimodal fusion to frame fire hazard recognition as a visual question answering task, enabling continuous fire safety monitoring with improved accuracy. Similarly, \citeN{adil2025using} developed a VLM-based framework for construction safety hazard identification, using prompt engineering to convert safety guidelines into contextual queries, allowing the model to analyze site images and generate hazard assessments aligned with regulations. Likewise, \citeN{zhang2025large} highlighted the potential of LLMs for CAD-based compliance checking, noting their ability to enhance efficiency, reduce manual effort, and support more accurate verification. A closely related work by \citeN{zhang2025doordet} leveraged LLMs and object detection to construct a functional multi-class door detection dataset, which required human-in-the-loop refinement. In contrast, our framework is fully automated, eliminating the need for human intervention.
\section{Methodology}
In this section, we first elaborate on our proposed new task for BCC, and then describe each component of our proposed pipeline in detail, one by one.
\subsection{Automated Facility Enumeration Task}
In this paper, we propose a new fundamental task for BCC, termed \textit{automated facility enumeration}, which is anticipated to become a core component of modern BCC in practice because enumeration of facilities is essential to the subsequent validation procedures for statutory requirements, such as NCC. Automated, as opposed to manual,  facility enumeration is targeted at large scale downstream validation. 

\begin{definition}[Automated Facility Enumeration]
Let $I$ denote a CAD floor plan image. The task of \textit{automated facility enumeration} 
is to automatically identify and count facilities of interest 
$\mathcal{F} = \{$sanitary facilities, kitchens, laundries, exits, 
fire safety equipment, accessibility provisions, parking$\}$ 
from $I$. 

Formally, the goal is to produce a mapping
\[
E: I \rightarrow \{(f_i, n_i)\}_{i=1}^{|\mathcal{F}|},
\]
where $f_i \in \mathcal{F}$ is a facility type and $n_i \in \mathbb{N}$ 
is the corresponding count. 
\end{definition}
\subsection{Door Detection}
\begin{figure}[htbp]
    \centering
    \includegraphics[width=\linewidth]{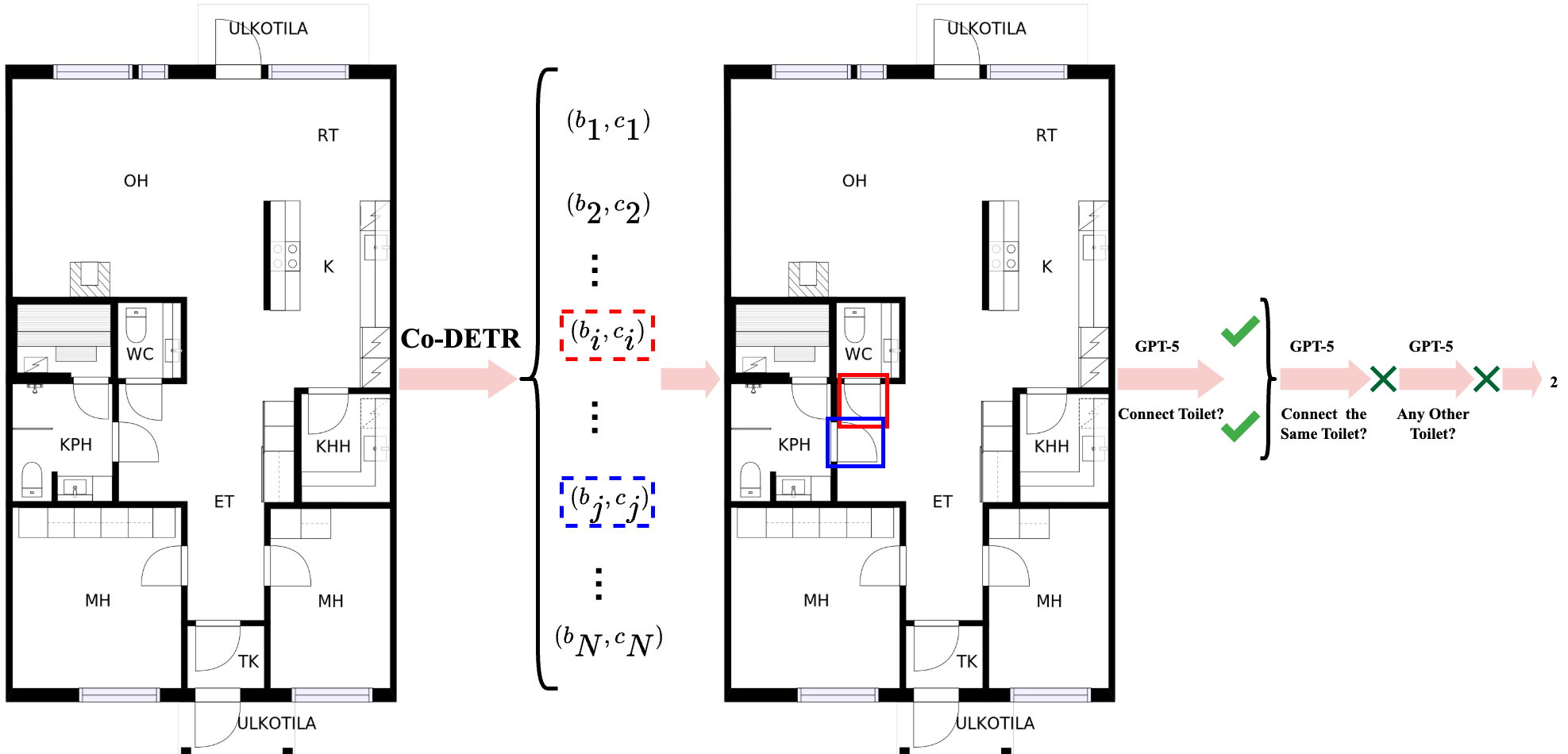}
    \caption{We illustrate our method using toilet prediction as an example. Given an input image, Co-DETR first detects all door instances. For each detected door, GPT-5 predicts whether it connects to a toilet. For all doors predicted to connect to a toilet, GPT-5 further determines whether they belong to the same toilet, removing duplicate instances. Finally, GPT-5 estimates whether any toilets have been missed. The image shown in the figure is from CubiCasa5K~\protect\cite{kalervo2019cubicasa5k}.
    \label{fig:method}}
\end{figure}

The complete workflow of our approach is illustrated in Figure~\ref{fig:method}. The process begins with door detection, serving as an essential precursor for accurate identification and analysis of facility types in the following stages.

Doors are present in most room designs. In our proposed method, door detection is a crucial step. To achieve optimal results and ensure robust generalization across domains, we adopt Co-DETR \cite{zong2023detrs}, a state-of-the-art object detection model benchmarked on COCO \cite{lin2014microsoft}. Its effectiveness for door detection has been validated in \cite{zhang2025doordet}. Co-DETR builds upon DETR \cite{carion2020end} by incorporating auxiliary heads trained with one-to-many label assignment, similar to Faster R-CNN \cite{ren2016faster}, in addition to the standard one-to-one matching of DETR. This design increases supervision during training by introducing more positive samples, which significantly improves detection performance.

Mathematically, the door detection step can be expressed as:
\begin{equation}
    \{(b_i, c_i)\}_{i=1}^N = \mathrm{Co\mbox{-}DETR}(\mathbf{I}),
\end{equation}
where $\mathbf{I}$ denotes the input floor plan image, $N$ is the number of detected instances, 
$b_i \in \mathbb{R}^4$ represents the predicted bounding box coordinates for the $i$-th door, 
and $c_i$ denotes its corresponding confidence score.

Door detection can achieve high accuracy when using state-of-the-art object detectors such as Co-DETR, particularly with abundant labeled training samples. Even in the presence of occasional missed or false detections, the CoT pipeline introduced in the following section is designed to tolerate such errors and still deliver strong overall performance.
\subsection{LLM Prediction via CoT}
In recent years, numerous LLM variants have emerged. In this paper, we adopt OpenAI’s latest flagship model, GPT-5 \cite{openai2025gpt5}, as the primary LLM in our framework. GPT-5 is a unified multimodal model that advances over prior versions in reasoning, factual accuracy, and adaptability across domains such as coding, mathematics, health, writing, and visual perception. Compared with earlier models, GPT-5 reduces hallucinations, improves instruction following, minimizes sycophancy, and provides more contextually nuanced, user-aligned responses.

To fully leverage the reasoning and multimodal capabilities of GPT-5, we propose a CoT pipeline tailored for the task. This pipeline decomposes the overall problem into a series of structured reasoning steps, allowing the model to methodically process and integrate visual and textual information.
\subsubsection{Room Connection Prediction}
Here, we use toilet prediction as a running example, while noting that the proposed method can be seamlessly extended to other facility types.  

Following door detection, the first step of the CoT pipeline is to predict the room connected to each detected door. For a given door, the LLM is prompted to determine whether it leads to a specific room type \(T\) (e.g., a toilet).

Formally, room connection prediction is defined as:
\begin{equation}
   P_i = \mathrm{LLM}\big(L_i, \mathbf{I} \mid \mathcal{P}\big),
\end{equation}
where $P_i \in \{\textit{yes}, \textit{no}\}$ denotes the prediction result for the $i$-th door, $L_i$ represents its location (given by the red bounding box in the floor plan), $\mathbf{I}$ is the input image, and $\mathcal{P}$ is the constructed prompt containing task instructions, room characteristics, and other contextual information.
\subsubsection{Room Consolidation}
After identifying all doors predicted to lead to a specific room type \(T\) (e.g., toilets), multiple doors may correspond to the same room. To resolve this, we use the LLM to perform pairwise room consolidation, determining which doors belong to the same room and eliminating duplicates to produce a single instance for each room.

Let \[\mathcal{D}_T = \{d_1, d_2, \dots, d_N\}\] denote the set of doors predicted to connect to room type \(T\). Many doors in \(\mathcal{D}_T\) may belong to the same room. To identify which doors correspond to the same room, we define a function that evaluates each pair of doors:
\begin{equation}
S_{ij} = \mathrm{LLM}\big(d_i, d_j, \mathbf{I} \mid \mathcal{P}_\text{room-assignment} \big),
\end{equation}
where $S_{ij} \in \{\textit{yes}, \textit{no}\}$ indicates whether doors $d_i$ and $d_j$ connect to the same room, $\mathbf{I}$ is the input image, $\mathcal{P}_\text{room-assignment}$ is the prompt designed to determine whether a given pair of doors connect to the same room.

After pairwise room consolidation, we obtain a deduplicated set of doors: 
\[
\mathcal{D}_T^\text{filtered} = \{d_1, d_2, \dots, d_M\}, \quad M \le N
\]
where $\mathcal{D}_T^\text{filtered}$ is the set of doors after pairwise room consolidation, containing only one representative door for each room. Here, $M$ is the number of doors remaining after consolidation, which is less than or equal to the original number of doors $N$.
\subsubsection{Omission Correction}
Following the previous steps, predictions are expected to be largely accurate. Nevertheless, some doors connecting to valid toilets may be misclassified by the LLM, leading to missing predictions. Additionally, open rooms without any associated doors are not detected or counted in the earlier steps. To address these gaps, we introduce a final step in the CoT pipeline specifically designed to capture any missing cases.

Specifically, all retained door predictions are highlighted in the image, and the LLM is prompted to detect any missing instances that lack a corresponding bounding box.

In formal terms,
\begin{equation}
N_\text{missing} = \mathrm{LLM}\big(\mathbf{I}, \mathcal{D}_T^\text{filtered} \mid \mathcal{P}_\text{missing}\big),
\end{equation}
where $N_\text{missing} \in \mathbb{Z}_{\ge 0}$ denotes the number of missing instances predicted by the LLM, $\mathbf{I}$ is the input image, and $\mathcal{P}_\text{missing}$ is the prompt instructing the LLM to identify and count instances not associated with a marked bounding box.

The final predicted number of a specific facility type is:
\[N_\text{final} = |\mathcal{D}_T^\text{filtered}| + N_\text{missing}.\]

Our proposed method is robust and generalizable across different datasets and scenarios, and it can be readily applied to any room type, including rooms without connected doors. Differences across room types are primarily captured in the prompts, which integrate both type-specific domain knowledge and the contextual scenario.
\section{Experiments}
\subsection{Datasets}
We select a diverse set of publicly available datasets to conduct our experiments. These datasets encompass various types of graphical documents, providing comprehensive coverage of symbol arrangements, document structures, and complexity levels. These subsets will be made publicly available to facilitate future comparisons.
\begin{enumerate}
    \item \textbf{CubiCasa5K}~\cite{kalervo2019cubicasa5k}. CubiCasa5K is a high-resolution floorplan dataset with annotations for over 80 object categories, providing rich semantic and geometric information for building layout understanding and architectural analysis. We use 97 test images for our experiments.
    \item \textbf{FloorPlanCAD}~\cite{fan2021floorplancad}. FloorPlanCAD is a large-scale dataset of real-world CAD floor plans from diverse building types, with expert annotations covering 30 object categories. We use 140 interior-layout samples from the test set for our experiments.
    \item \textbf{BRIDGE}~\cite{goyal2019bridge}. BRIDGE is a large-scale dataset of floor plan images collected from public sources, with annotations for 16 classes of interior symbols, room regions, and textual descriptions of complete layouts. We use 113 high-quality samples for our experiments.
    \item \textbf{SESYD}~\cite{delalandre2010generation}. SESYD is a synthetic dataset of 1000 document images, including architectural floorplans and electrical diagrams, containing over 57000 annotated symbols across diverse backgrounds. We use 48 images from this dataset for our experiments.
    \item \textbf{ROBIN}~\cite{sharma2017daniel}. ROBIN is a dataset of 510 real-world floor plans annotated using standard notations, capturing diverse room types and layouts. We use 48 images from this dataset for our experiments.
    \item \textbf{SydneyHouse}~\cite{chu2016housecraft}. SydneyHouse is a dataset of 174 residential houses across Sydney, Australia, with detailed floorplans, approximate geo-locations, and wide-baseline StreetView images capturing the building exteriors. We use 88 images from this dataset for our experiments.
    \item \textbf{CVC-FP}~\cite{de2015cvc}. CVC-FP is a dataset of real architectural floor plan images, supporting high-resolution files and annotated with key elements and their spatial and functional relationships. Only a resized version ($640\times640$) is available, so we select 15 relatively clear images for our experiments.
    \item \textbf{MLSTRUCT-FP}~\cite{pizarro2023large}. MLSTRUCT-FP contains 954 high-resolution multi-unit floor plan images from Chilean residential buildings. For our experiments, we select 6 samples (345 door instances). Each image is divided into several parts for processing, and the results from all parts are combined to obtain the final prediction.
\end{enumerate}
\subsection{Evaluation Metric and Baseline}
To validate our proposed method, we use accuracy as the evaluation metric, comparing the predicted count of each facility type against the ground truth. An image is considered correctly predicted if the number of instances for a given type is estimated accurately.

We also use the LLM’s predictions as a baseline by prompting it to estimate instance counts. A prediction is considered accurate if the estimated count matches the ground truth, regardless of spatial positions. While the LLM can describe object locations, the exact positions remain ambiguous. Similarly, our method is deemed accurate when the predicted count for a specific type matches the ground truth.
\subsection{Experiment Settings}
To train the door detection model, we used two publicly available datasets \cite{floorplans500,doorobjectdetection}. The first \cite{doorobjectdetection} contains 1047 training and 62 test samples, and the second \cite{floorplans500} contains 837 training and 43 test samples. The training samples were combined into a unified training set, and the test samples were merged into a single test set. To further improve accuracy, we manually labeled a subset of examples from each dataset, which will be made publicly available in the future.
\subsection{Experimental Results}
\noindent\textbf{Results for the Toilet Category:} We begin by evaluating the toilet category. Table \ref{tab:toilet} presents a comparison between our proposed CoT pipeline and the GPT-5 baseline. Across all eight experiments, our method consistently outperforms the baseline, demonstrating the effectiveness of the CoT approach in capturing complex spatial and contextual cues for accurate facility prediction.
\begin{table}[t]
\centering
\caption{Comparison of the baseline and our CoT-based method for the toilet category across eight publicly available datasets.}
\begin{tabular}{c|c|c}
\toprule
 \multirow{2}{*}{\textbf{Datasets}} & \multicolumn{2}{c}{\textbf{Accuracy (\%)}} \\
 \cmidrule{2-3}
 & GPT-5 & \textbf{Ours} \\
 \midrule
 CubiCasa5K~\cite{kalervo2019cubicasa5k} & 85.57 & \textbf{87.63} \\
 \midrule
 FloorPlanCAD~\cite{fan2021floorplancad} & 79.29 & \textbf{85.71} \\
 \midrule
 BRIDGE~\cite{goyal2019bridge} & 77.88 & \textbf{81.42} \\
 \midrule
 SESYD~\cite{delalandre2010generation} & 68.75 & \textbf{95.83} \\
 \midrule
 ROBIN~\cite{sharma2017daniel} & 64.58 & \textbf{85.42} \\
 \midrule
 SydneyHouse~\cite{chu2016housecraft} & 57.95 & \textbf{71.59} \\
 \midrule
 CVC-FP~\cite{de2015cvc} & 40.00 & \textbf{86.67}\\
 \midrule
 MLSTRUCT-FP~\cite{pizarro2023large} & 50.00 & \textbf{66.67} \\
 \bottomrule
\end{tabular}
\label{tab:toilet}
\end{table}

\noindent\textbf{Results for the Kitchen Category:} We next evaluate our method on the kitchen category. Table~\ref{tab:kitchen} reports the comparative results between our CoT-based approach and the GPT-5 baseline on the CubiCasa5K and SydneyHouse datasets. In both experiments, although the baseline performs strongly, our method achieves slightly better results, indicating the superiority of the proposed approach.

\begin{table}[t]
\centering
\caption{Comparison of the baseline and our CoT-based method for the kitchen category using two selected publicly available datasets.}
\begin{tabular}{c|c|c}
\toprule
 \multirow{2}{*}{\textbf{Datasets}} & \multicolumn{2}{c}{\textbf{Accuracy (\%)}} \\
 \cmidrule{2-3}
 & GPT-5 & \textbf{Ours} \\
 \midrule
 CubiCasa5K~\cite{kalervo2019cubicasa5k} & 96.91 & \textbf{98.97} \\
 \midrule
 SydneyHouse~\cite{chu2016housecraft} & 97.73 & \textbf{98.86} \\
 \bottomrule
\end{tabular}
\label{tab:kitchen}
\end{table}

\noindent\textbf{Results for the Emergency Exit Door Category:}  
We further evaluate our method on the emergency exit door category. According to the NCC, exit doors are required to ensure safe evacuation, while no notion of an `emergency room' is defined. Therefore, the task is formulated as verifying the number of doors serving as emergency exit doors. We compare our proposed method with GPT-5 on the CubiCasa5K dataset. Table~\ref{tab:emergencyexit} presents the results. As shown, our CoT-based method consistently outperforms the baseline GPT-5, demonstrating both the necessity and effectiveness of incorporating CoT reasoning.

\begin{table}[t]
\centering
\caption{Comparison of the baseline and our CoT-based method for the emergency exit category on the CubiCasa5K dataset.}
\begin{tabular}{c|c|c}
\toprule
 \multirow{2}{*}{\textbf{Datasets}} & \multicolumn{2}{c}{\textbf{Accuracy (\%)}} \\
 \cmidrule{2-3}
 & GPT-5 & \textbf{Ours} \\
 \midrule
 CubiCasa5K~\cite{kalervo2019cubicasa5k} & 67.01 & \textbf{70.10} \\
 \bottomrule
\end{tabular}
\label{tab:emergencyexit}
\end{table}

\noindent\textbf{Results for the Car Parking Category:} Finally, we evaluate the performance of our CoT method and the baseline on the car parking category using the CubiCasa5K and SydneyHouse datasets. Table~\ref{tab:carpark} summarizes the quantitative results for each dataset. On CubiCasa5K, our CoT method performs slightly better than GPT-5, while on SydneyHouse, our method outperforms the baseline by a large margin, clearly demonstrating the advantage of integrating CoT reasoning with LLMs.

\begin{table}[t]
\centering
\caption{Comparison of the baseline and our CoT-based method for the car parking category using two selected publicly available datasets.}
\begin{tabular}{c|c|c}
\toprule
 \multirow{2}{*}{\textbf{Datasets}} & \multicolumn{2}{c}{\textbf{Accuracy (\%)}} \\
 \cmidrule{2-3}
 & GPT-5 & \textbf{Ours} \\
 \midrule
 CubiCasa5K~\cite{kalervo2019cubicasa5k} & 96.91 & \textbf{97.94} \\
 \midrule
 SydneyHouse~\cite{chu2016housecraft} & 87.50 & \textbf{93.18} \\
 \bottomrule
\end{tabular}
\label{tab:carpark}
\end{table}

Overall, the experiments confirm the critical role and practical advantage of the proposed CoT pipeline in enabling the adopted LLM method to perform facility enumeration in a generalizable, robust, and accurate manner, which is vital for meeting BCC requirements under the NCC.

Notably, through experiments, we found that appending ``tell me the reason'' at the end of the prompt consistently improves performance. It suggests that encouraging the LLM to explicitly justify its predictions fosters more deliberate reasoning, thereby enhancing accuracy.
\section{Conclusion}
In this paper, we address the challenges of accurate facility enumeration for compliance verification by focusing on BCC requirements under the NCC and exploring how these regulatory rules can be translated into automated analysis tasks. Based on the identified compliance rules, we introduce a new task, automated facility enumeration, which systematically identifies and counts key facilities in floor plans, including sanitary facilities, kitchens, laundries, exits, fire safety equipment, accessibility features, and parking. To address this task, we leverage the reasoning capabilities of LLMs. Furthermore, we propose a CoT framework that integrates door detection with LLM-based reasoning to enhance precision and robustness. Extensive experiments across diverse datasets and facility types demonstrate that our approach achieves accurate and generalizable results, highlighting its potential to support intelligent, scalable compliance checking workflows.
%
%
\bibliography{ascexmpl-new}

\end{document}